# North Atlantic Right Whale Contact Call Detection

**Rami Abousleiman**                                                                 RDABOUSL@OAKLAND.EDU
Oakland University, Department of Electrical and Computer Engineering, Rochester, MI, USA

**Guangzhi Qu**                                                                      GQU@OAKLAND.EDU
Oakland University, Department of Computer Science and Engineering, Rochester, MI, USA

**Osamah Rawashdeh**                                                                 RAWASHD2@OAKLAND.EDU
Oakland University, Department of Electrical and Computer Engineering, Rochester, MI, USA

## Abstract

The North Atlantic right whale (Eubalaena glacialis) is an endangered species. These whales continuously suffer from deadly vessel impacts alongside the eastern coast of North America. There have been countless efforts to save the remaining 350 - 400 of them. One of the most prominent works is done by Marinexplore and Cornell University. A system of hydrophones linked to satellite connected-buoys has been deployed in the whales' habitat. These hydrophones record and transmit live sounds to a base station. These recording might contain the right whale contact call as well as many other noises. The noise rate increases rapidly in vessel-busy areas such as by the Boston harbor. This paper presents and studies the problem of detecting the North Atlantic right whale contact call with the presence of noise and other marine life sounds. A novel algorithm was developed to preprocess the sound waves before a tree based hierarchical classifier is used to classify the data and provide a score. The developed model was trained with 30,000 data points made available through the Cornell University Whale Detection Challenge program. Results showed that the developed algorithm had close to 85% success rate in detecting the presence of the North Atlantic right whale.

## 1. Introduction

The dependency on ship transportation for goods has increased the ocean congestion alongside the eastern side of the United States and Canada. The North Atlantic right whale (NARW) suffers from this increase Jensen et al. (2004). NARW is a mammal and thus requires air to breath. Heading towards the ocean surface can be dangerous as impacts with one of these large vessels may be deadly. Ship crews seldom notice the presence of the whale(s) and most times there is nothing to do after a hit has occurred. This problem is further escalated due to the fact the NARW is an endangered species with only a couple hundred of these mammals survive nowadays Kraus et al. (2005), Caswell et al. (1999), and Fujiwara et al. (2001).

Countless efforts have been done to conserve and study the behavior of the NARW Matthews et al. (2001), Clark et al. (2007), Vanderlaan et al. (2003), and Parks et al. (2005). To help solve this problem an autonomous near-real-time buoy system for automatic detection of NARW has been developed Spaulding et al. (2009). The NARW makes a unique call known as 'contact call' or 'up-call' these calls are used as a communication method between the whales as a way of letting each other know of their presence. This distinguished call has unique characteristics that will be used as the benchmark for the NARW call detection. This paper proposes a method to analyze the sound recording acquired by the deployed buoys Spaulding et al. (2009) and then describe the developed algorithm that will automate the detection process.

The rest of the paper is organized as follows: Section 2 will present the related work. The characteristics of the up-call with emphasis on the significant features will follow in the Section 3. Details on the implemented algorithm with its parameters and flowchart will then be described followed by the results and discussion. The paper is then ended with a future work section and a conclusion.

## 2. Related Work

Many work has been done to detect the presence of certain animal species based on their sounds or calls. Whales and birds are mostly studied because of their unique communication capabilities. Some studies were done to classify birds. These studies help in reducing bird strikes with airplanes especially near airports Kwan et al. (2006). The authors used a Hidden Markov Model and Gaussian Mixture Models to do suitable real-time monitoring for a large number of birds. Marcarini et al. (2008) studied specific kinds of migrating birds using Gaussian Mixture Models alongside with spectrogram correlation. In fact, many studies use spectrogram correlation when it comes to acoustic analysis Mellinger et al. (2000), Chabot (1988), and Mellinger et al. (2000). The study of the NARW call is not that common. The rest of this section will specifically discuss the NARW contact call classification. Dugan et al. (2010) used classification and regression trees

(CART) and artificial neural networks (ANN) methods. The results were then compared with a feature vector testing (FVT) approach. The CART had the highest assignment rate. The FVT had low false positive rates, however, it also had an overall assignment rates less than the ANN method. Dugan et al. (2010) improved on the results that were generated in their previous work. The proposed method features multiple algorithms running in parallel. The output of the individual algorithms is then fed into a decision algorithm that provides the final output of the system. The developed algorithm had a better detection probability than the FVT. Side by side comparison between the FVT and the developed method showed that the developed method had lower number of false positive rates. The authors however did not mention the performance when it comes to missing a positive call. Urazghildiiev el al. (2010) developed a multistage, hypothesis testing technique that involves the generalized likelihood test detector. The proposed algorithm was able to detect approximate 85% of the contact calls that were detected by a human operator. The algorithm had about 26 false alarms per day. The method implemented in this paper uses a novel algorithm that preprocesses the sound waves before they are passed into a tree based classifier to output the final results. The next section will discuss the distinctive features that uniquely identify the contact call of the NARW.

## 3. Contact Call Features

The unique characteristics of the NARW are the key components that help identify the presence of a NARW even in the existence of noise as well as in the presence of sounds from other kinds of whales. Figure 1 shows a sample NARW contact call. The duration, minimum frequency, maximum frequency, and bandwidth are some of the features that uniquely distinguish the NARW from a pool of sounds. For example, 99% of these calls are within 0.3 to 1.5 seconds in range Gillespie et al. (2004). Another key feature of the signal is that the upsweep part comprises of 30 to 100% of the total signal duration. A list of features was identified by Dugan et al. (2010). Table 1 lists these features; ultimately these features were used to classify the NARW up call.

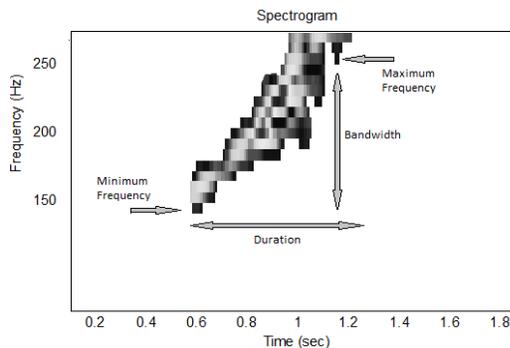

*Figure 1.* Sample North Atlantic Right Whale Contact Call

*Table 1.* List of Features used to Classify the North Atlantic Right Whale Up-Calls

| Feature | Description |
|---------|-------------|
| $f_1$ | Signal Duration |
| $f_2$ | Minimum Frequency |
| $f_3$ | Maximum Bandwidth |
| $f_4$ | Start-end Bandwidth |
| $f_5$ | Duration of Upsweep |
| $f_6$ | Local Noise Level |
| $f_7$ | Segmentation thresholds |
| $f_8$ | Mean Value of the Instantaneous Bandwidth |
| $f_9$ | Percentage of holes in the object |
| $f_{10}$ | Percentage of down sweeps in object |
| $f_{11}$ | Percentage of harmonics in the object |

## 4. Developed Algorithm

Given the features and the signal properties provided in the previous section, a procedure was developed to pre-process the sound signals and a novel algorithm was proposed to determine whether a NARW is present or not. The sound signal pre-analysis will be discussed in the next section followed by the description and analysis on the implemented algorithm.

### 4.1 Sound Signal Pre-Analysis

Before the developed algorithm can be implemented a pre-analysis is done on the sound signals. Pre-analysis constitutes of three steps listed below:

1. Read the sound wave and then use fast fourier transform to convert the signal from the time domain to the frequency domian
2. Ignore the weakest 80% of the sound array
3. Clear data islands using the weakest neighborhood points

Step 2 above is implemented as follows: The sound array is sorted in ascending order and the smallest 80% of the sound vector are eliminated and replaced by zero. The weakest neighborhood method is a novel algorithm that aims at clearing data islands. It does not eliminate 100% of the data islands but it does significant improvements on the data matrix and makes it easier to handle. The definition of data islands will get clear in the next section. Algorithm 1 shows the weakest neighborhood method.

Lines 1 to 10 in Algorithm 1 show the main method. Lines 11 to 32 illustrate the *clearDataIslands* function that is called in line 4. Lines 1 and 11 initiate the terminate conditions for the routine and the subroutine respectively. The routine in lines 1 to 10 keeps calling the *clearDataIslands* function until there is no improvement on the results. Another termination criterion might also be added in case there is a computational concern. For example, a limit might be set on the number of improvements rather than waiting until no improvement is possible. The subroutine *clearDataIslands* gets the neighborhood of each non-zero non-edge point in the

*DataMatrix* and then finds if less than 4 of the neighboring points has a value of 0 (line 24) if this is the case then that specific point is replaced with zero otherwise it is left unchanged. The subroutine is terminated when all the points in the *DataMatrix* have been tested. The results of the signal preprocessing section are significant and it sets the path towards easier processing in the next section. Figure 3 shows 2 examples of different sound recording after the implementation progresses through the 3 steps described earlier.

**Algorithm 1** Weakest Neighborhood Method
**Input:** *DataMatrix*
**Output:** *DataMatrix$_{MODIFIED}$*
1 StopCondition1 ← 0
2 **while** ¬ StopCondition1 **do**
3     temp = DataMatrix
4     DataMatrix = clearDataIslands(DataMatrix)
5     **if** temp = DataMatrix **then**
6       DataMatrix$_{MODIFIED}$ = temp
7       StopCondition1 = TRUE
8     **end**
9 **end**
10 **return** DataMatrix$_{MODIFIED}$

*DataMatrix* = clearDataIslands(*DataMatrix*)
11 StopCondition2 ← 0
12 **while** ¬ StopCondition2 **do**
13     ∀ *non* − *edge point* $i \neq 0 \in DataMatrix$ **do**
14     [$a\ b\ c\ d\ e\ f\ g\ h$] = $getNeigherhood(i)$
15     **if** a >0 **then** a=1 **end**
16     **if** b >0 **then** b=1 **end**
17     **if** c >0 **then** c=1 **end**
18     **if** d >0 **then** d=1 **end**
19     **if** e >0 **then** e=1 **end**
20     **if** f >0 **then** f=1 **end**
21     **if** a >0 **then** g=1 **end**
22     **if** a >0 **then** h=1 **end**
23     $SUM = a + b + c + d + e + f + g + h$
24     **if** SUM < 4
25       update *DataMatrix* by setting
26       (*non* − *edge point* $i$ = 0)
27     **end**
28     **if** ∄ *non* − *edge point*
29       StopCondition2 = TRUE
30     **end**
31 **end**
32 **return** *DataMatrix*

Figure 2 shows the return parameters of the $getNeigherhood(i)$ function that is called in line 14 of Algorithm 1.

| a | b | c |
|---|---|---|
| d | *i* | h |
| e | f | g |

*Figure 2.* The Return Parameters of the $getNeigherhood$ Function

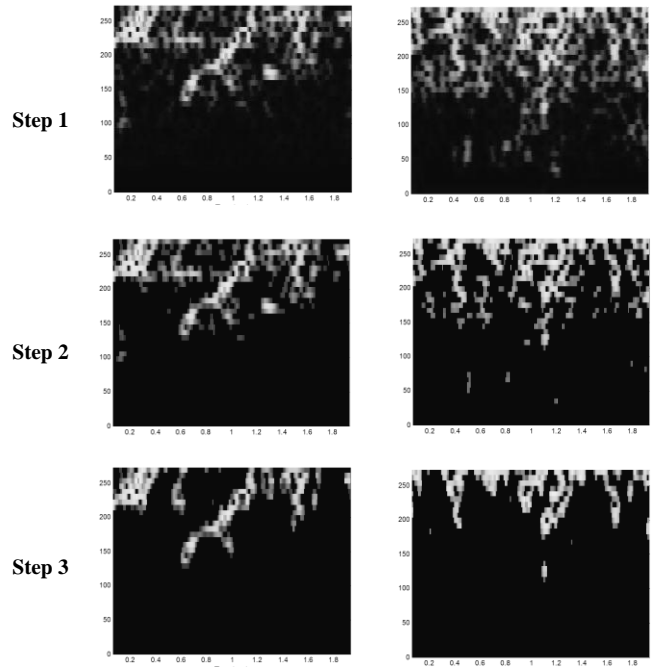

Step 1

Step 2

Step 3

*Figure 3.* Spectogram Changes after Implementing Steps 1, 2 and 3 as Described in the Signal Pre-Analysis Section

### 4.2 NARW Detection Algorithm

The output of the pre-processing of the sound signals is then used as the input to the developed algorithm. The developed algorithm consists of 8 steps. These steps are listed in their sequence of execution in the following order:

1. Initiate 10 equally spaced particles $p_1, p_2, \ldots, p_{10}$ on the time axis (For example, if the sound wave spectrogram is 2 seconds in length then the particles will be located at [0.00 0.22 0.44 0.66 0.88 1.11 1.33 1.55 1.77 2.00]).
2. These particles are set to explore and determine the first non-zero terms on the frequency matrix (Figure 4 shows the selected particles for a sample sound recording example)

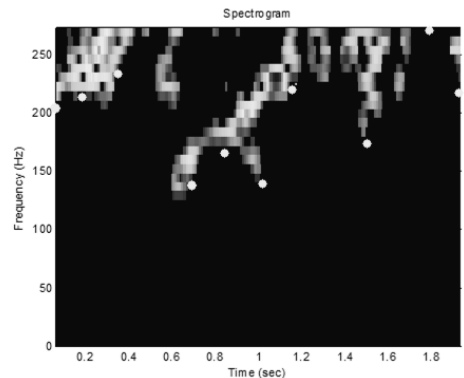

*Figure 4.* Circles Representing the Particles (Step 2)

3. Particle $p_1$ then travels to particle $p_2$, particle $p_2$ travels to particle $p_3$, etc…The particles travel only on non-zero paths (i.e. when the frequency > 0). The equation of motion is expressed as follows: Roulette wheel selection is used to pick the next point of motion. The probability of picking a point $j$ in the neighborhood of $i$ is given by a deterministic factor and a random factor. The equation that determines that factors is expressed by Equation (1)

$$\alpha \left(\frac{1}{1+(f(p_n)-\mu)^2}\right) + \beta x \qquad (1)$$

Where $\alpha$ and $\beta$ are the learning parameters ($\alpha + \beta = 1$). $x$ is a random number with the range from 0 to 1. $f(p_n)$ is the frequency value at point $p_n$. $\mu$ is the mean value (of the frequency values) between $p_n$ and $p_{n+1}$. Figure 5 shows the result of executing step 4. If there is no route between the points then the path is not generated and it will be eliminated in the next step.

Equation 1 is inspired by the Particle Swarm Optimization technique and by how the particles move from a random point to the minimum or maximum in an optimization problem Kennedy et al. (1995) and Venter et al. (2003). The values of $\alpha$ and $\beta$ can be adjusted to favor the deterministic factor versus the stochastic factor, if $\beta$ increases then $\alpha$ must decrease thus favoring the second term in (1). For example, setting $\alpha$ to 0 will cause the motion between $p_n$ and $p_{n+1}$ to be completely stochastic.

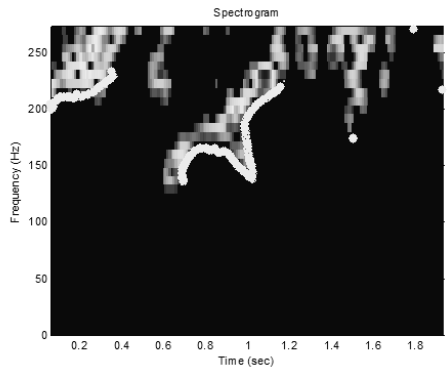

Figure 5. The Result of Executing Step 4

4. All found paths that are < 0.3 seconds are ignored (The 0.3 value came from the fact that no NARW contact call should be less than 0.3 seconds in duration). Figure 6 shows the outcome of this step.
5. Repeat step 1 (initiate 10 points but this time on the new time axis as shown in Figure 7)

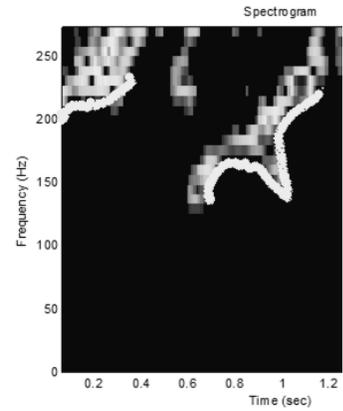

Figure 6. Outcome of Step 4

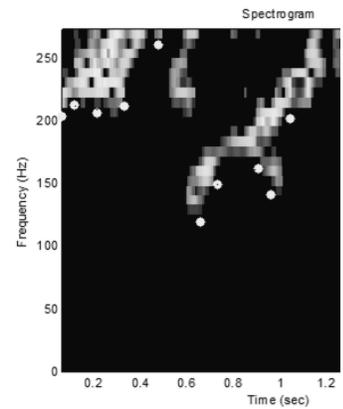

Figure 7. Outcome of Step 5

6. Repeat step 3 (particles travel from one point to the other) until there are no paths that are <0.3 seconds in length. Figure 8 shows the outcome of step 6.

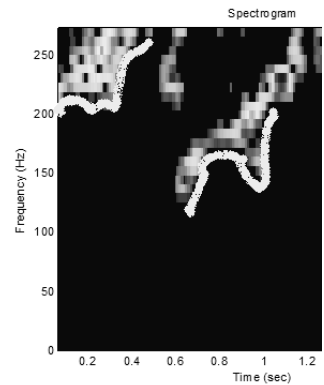

Figure 8. Outcome of Step 6

7. Let the particles propagate in the y-axis (frequency) until a zero element is observed. Figure 9 shows the outcome of this step.
8. Any particle that fails in observing a zero element shall be ignored. Furthermore any isolated particle

shall also be ignored. Figure 10 shows the outcome of step 8.

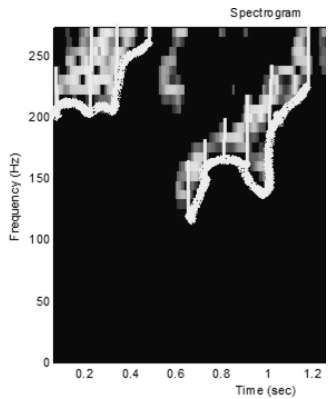

Figure 9. Outcome of Step 7

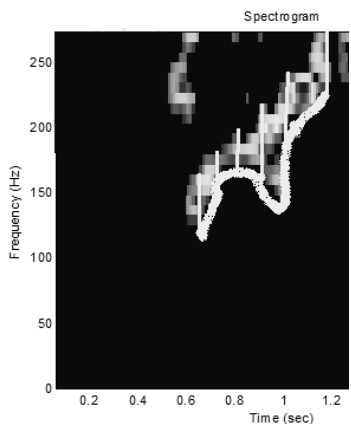

Figure 10. Outcome of Step 8

### 4.3 Feature Recognition Process

The feature recognition process starts with the output that was generated from step 8 from the previous section. All the found path(s) are used in the feature recognition process. The features listed in Table I are identified and a score is given to the path(s). Gaussian based assignment is used for this process. For example, the length of the path ($f_1$ in Table I) should be between 0.3 seconds to 2 seconds. On the other hand, 99% of the signal lengths are between 0.3 to 1.5 seconds. Thus it makes more sense to give more score to the signals that fall within this range. Figure 11 shows an example of how would the score be distributed for $f_1$. The same is done to all of the features listed in Table 1. The Gaussian mean and variance, that defines how the curve deviates, change for every feature according to the trained data. Once the score for every feature is calculated, these scores will be used as the split decision factor in the tree based classifier until a decision is made.

## 5. Results

The algorithm was trained with a sample of 30,000 points. The success rate was at 84.7%. For a sample of 1000 sound signals, 112 were identified as false positive calls. On the other hand, 41 calls were mistakenly identified as a non NARW contact call.

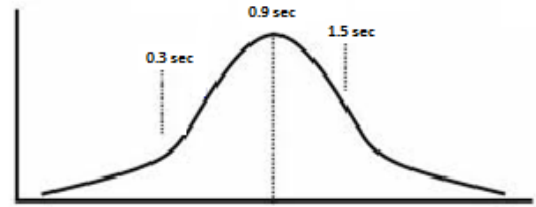

Figure 11. Gaussian Distribution for Feature $f_1$

## 6. Future Work

Many improvements can be done on the designed algorithm to enhance the results. Parameters $\alpha$ and $\beta$ can be better estimated to provide the optimal combination between the deterministic versus the stochastic factor. Another improvement is to optimally design the Gaussian based assignment described earlier. The design should adhere to each of the specific feature vectors. The weakest neighborhood method can also be enhanced to have better coverage by deleting the unnecessary data islands that might still exist after the method had executed. Finally, the algorithm should improve, to guarantee a no-miss of the whale up-call. Although the miss rate is low but getting the number closer to zero will have a huge improvement on the solution.

## 7. Conclusion

In this paper, an algorithm was presented to detect the presence of an up-call of the North Atlantic right whale. 30,000 recording were used to train the model that was based on a tree classifier. The algorithm proved to successfully work by detecting the contact calls with a success rate close to 85%.